%% file: ms.tex
\pdfoutput=1

\documentclass[times]{article}

\usepackage[final,nonatbib]{nips_2018} 

\input{setup}

\title{Improving Clinical Predictions through Unsupervised Time Series Representation Learning}

\author{
  Xinrui Lyu$^1$, Matthias H\"user$^1$, Stephanie L. Hyland$^1$, George Zerveas$^2$, Gunnar R\"atsch$^1$\\
  1 Biomedical Informatics Group, Dept. of Computer Science, ETH Z\"urich \\
  2 AI Lab, Center for Biomedical Informatics, Brown University \\
}

\begin{document}

\maketitle

\input{abstract}
\input{introduction}
\input{related_work}
\input{models}
\input{experiments}

\input{conclusion}

\small

\bibliographystyle{unsrt}
\bibliography{ms}

\clearpage
\input{appendix}
\end{document}

%% file: setup.tex
\usepackage[footnotesize]{caption}
\usepackage{picinpar}
\usepackage[utf8]{inputenc}
\usepackage[T1]{fontenc}        
\usepackage{url}           
\usepackage{booktabs}       
\usepackage{amsfonts}      
\usepackage{nicefrac}       
\usepackage{microtype}    
\usepackage{array}
\usepackage[table]{xcolor}
\usepackage{placeins}
\usepackage{pgfplots}
\usepackage{amsmath}
\usepackage{amssymb}
\usepackage[sort&compress,numbers]{natbib}
\usepackage{graphicx}           
\usepackage{subcaption}
\usepackage{multicol}
\usepackage{placeins}
\usepackage{multirow}
\usepgfplotslibrary{colorbrewer}
\pgfplotsset{cycle list/Dark2-8}
\usepackage{numprint}

%% file: abstract.tex
\begin{abstract}
In this work, we investigate unsupervised representation learning on medical time series, which bears the promise of leveraging copious amounts of existing unlabeled data in order to eventually assist clinical decision making. By evaluating on the prediction of clinically relevant outcomes, we show that in a practical setting, unsupervised representation learning can offer clear performance benefits over end-to-end supervised architectures. We experiment with using sequence-to-sequence (Seq2Seq) models in two different ways, as an autoencoder and as a forecaster, and show that the best performance is achieved by a forecasting Seq2Seq model with an integrated attention mechanism, proposed here for the first time in the setting of unsupervised learning for medical time series.
\end{abstract}

%% file: introduction.tex
\vspace{-0.2cm}
\section{Introduction}
\vspace{-0.2cm}

Patient representation learning is one of the popular topics in the field of machine learning for healthcare.
The generality of supervised representations is usually constrained by the amount of labeled data, while unsupervised representations can leverage information from all data, labeled or not.
Hence, unsupervised learning can produce representations of \emph{general} utility~\cite{dosovitskiy2014unsupervisedForImages, mikolov14doc2vec,mikolov13word2vec,miotto16_deep_patient_unsupervised}, which can be useful in case downstream tasks are not known \emph{a priori}.

Conditions like the ones described above are especially true in the medical domain. Routine medical practice generates a wealth of patient-related time series, while data annotation often requires medical experts, whose time is very limited. Additionally, new tasks of interest emerge, and different hospitals or health systems often define tasks in different ways. Thus, generally useful representations, providing good performance over a broad range of downstream tasks, are highly desired.\

In this work, we investigate unsupervised representation learning on medical time series, which remains relatively unexplored. We propose adapted and novel models well suited for this objective and elucidate under which conditions they provide a performance benefit over end-to-end supervised learning with respect to predicting clinically relevant outcomes.

%% file: related_work.tex
\vspace{-0.2cm}
\section{Related Work} 
\vspace{-0.2cm}

The unsupervised learning approaches studied in this paper are rooted in the autoencoding principle~\cite{bengio2013representation}.
The basic autoencoding architecture has been extended in several ways, such as denoising \cite{vincent2010stacked}, variational \cite{kingma2013auto}, convolutional \cite{masci2011stacked}, or contractive \cite{rifai2011contractive} autoencoders. Sequence-to-sequence (Seq2Seq)~\cite{sutskever2014sequence} architectures have been used successfully in translation~\cite{weiss2017sequence}, and on text and images~\cite{chen2015mind,gregor2015draw}. Seq2Seq models have also been pre-trained in an unsupervised way \cite{ramachandran2016unsupervised} and fine-tuned with labeled data.

Several models for unsupervised representation learning have been successfully employed in medical applications~\cite{pivovarov2015learning,miotto16_deep_patient_unsupervised,suresh17_use_autoencoders_discovering,jones16_canonical_correlation_analysis,choi2016multi}. 
While in many cases representations were obtained with both descriptive as well as predictive utility, the optimal reconstruction principles and loss functions leading to accurate clinical outcome prediction have not been widely studied.

Attention mechanisms can improve performance and interpretability and have enjoyed wide use across domains~\cite{chorowski2015attention,xu2015show,kumar2016ask,choi2016retain}. 
Although attention has been used in the context of unsupervised representation learning of natural language \cite{jang2018RNNSVAE}, attention architectures in the medical domain  have been so far exclusively focused on predicting specific supervised tasks.

%% file: models.tex
\vspace{-0.2cm}
\section{Representation Learning Models}\label{sec:section}
\vspace{-0.2cm}
\subsection{Baselines: Autoencoders}

Autoencoding consists of two steps: \textit{encoding} maps the input data space $\mathbb{R}^d$ to an representation space $\mathbb{R}^m$, where typically $m<d$, while \textit{decoding} maps in the reverse direction to reconstruct the data from representations.
The objective of autoencoding is to minimize the reconstruction error between the input data and the reconstructions.

\vspace{-0.1cm}
\paragraph{Non-Sequential models}
Principal Component analysis (PCA) and its inverse together can be considered as a simple autoencoding process, where the encoding is a learned linear projection.
An autoencoder (AE) is a neural network composed of an encoder and a decoder, each implemented as a multi-layer perceptron; it encodes the data in a non-linear way.
Our goal is to encode temporal sequences of physiological signal vectors, but the inherent architecture of PCA and AE does not allow them to exploit the temporal structure in time series.
To make data compatible with the input format of PCA and AE, we flatten a $(T, d)$-dimensional time series (i.e. $T$ time samples, each of $d$ dimensions) into a $(T d)$-dimensional vector.

\begin{figure}[!b]
\begin{center}\vspace*{-3.5ex}\includegraphics[width=0.75\textwidth,trim={0 6.5cm 0 2.7cm},clip]{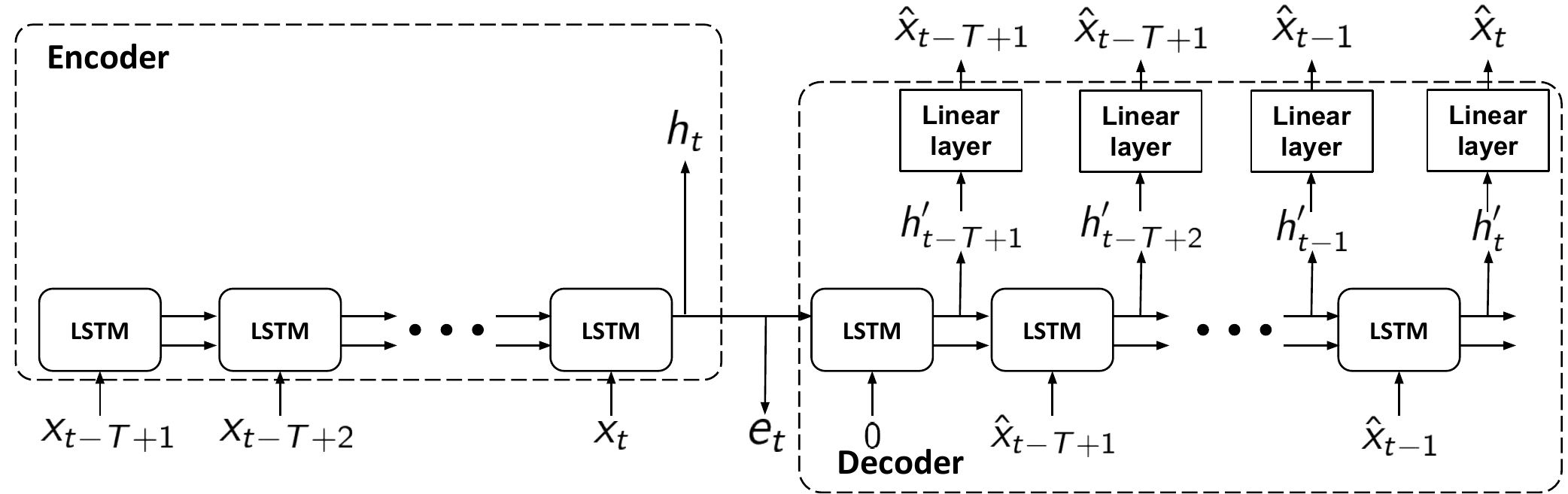}\vspace*{-2ex}\end{center}
\caption{Sequence-to-sequence autoencoder (S2S-AE). $T$ is the length of history the we want to encode in the representations. If we want to encode the patient's history from admission to time $t$, then $T=t$.\vspace*{-3.5ex}}
\label{fig:new_Seq2Seq}
\end{figure}

\vspace{-0.1cm}
\paragraph{Seq2Seq model}
While Seq2Seq models are often used in supervised training settings in natural language processing~\cite{sutskever2014sequence,ramachandran2016unsupervised,weiss2017sequence}, we use it in an unsupervised way by minimizing the input reconstruction error as an objective; we refer to such a model as a S2S-AE.
Figure~\ref{fig:new_Seq2Seq} shows the structure of a S2S-AE model. A Long Short-Term Memory (LSTM) cell is used for both encoder and decoder recurrent neural network (RNN) units, because it can retain information over more time-steps compared to simple RNN cells \cite{hochreiter1998vanishing,hochreiter2001gradient}.

At time $t$, the encoder receives a sequence of signal vectors $X_t:=\{\mathbf{x}_{\tau}\}_{\tau=t-T+1}^{t}$ from a time window of size $T$ as input and produces a representation $\mathbf{e}_t := h_t$, where $h_t$ is the last hidden state of the encoder.
The decoder, given $\mathbf{e}_t$, outputs a sequence of reconstructions $\{\hat{\mathbf{x}}_{\tau}\}_{\tau=t-T+1}^{t}$ for the same window.
Let $f_\theta$ and $g_\phi$ denote the encoder and decoder respectively, with parameters $\theta$ and $\phi$. Then the S2S-AE model can be formulated like 
\begin{align}
 \label{eq:s2s-c}f_\mathbf{\theta}(\mathbf{x}_{t-T+1}, \mathbf{x}_{t-T+2}, \ldots, \mathbf{x}_{t}) = \mathbf{e}_{t} ,~~&~~~g_\mathbf{\phi}(\mathbf{e}_{t}) = \hat{\mathbf{x}}_{t-T+1}, \hat{\mathbf{x}}_{t-T+2}, \ldots, \hat{\mathbf{x}}_{t},\\
 \label{eq:s2s-c_obj}\mathcal{L}(X_t) = \frac{1}{T}\sum\nolimits_{\tau=0}^{T-1} \| \mathbf{x}_{t-\tau} - \hat{\mathbf{x}}_{t-\tau}\|^2,~~&~~~
 \mathcal{L} = \frac{1}{N}\sum\nolimits_{i=1}^{N}\frac{1}{L_i}\sum\nolimits_{k=0}^{L_i-1}\mathcal{L}(X_{T+k}),~~
 \end{align}
where $\mathcal{L}(X_t)$ is the average reconstruction error for one window of a single patient's input signals from $t-T+1$ until the current time $t$.
The loss for patient $i$ is then the average error over their $L_i$ windows, indexed by $k$, sliding with stride 1. To train the S2S-AE model we average the patient-wise loss over all $N$ patients. 
The representation $\mathbf{e}_t$ from a S2S-AE model summarizes a fixed length of the medical history of a patient up to time $T$, which reflects the current state of the patient.

\subsection{Sequential forecasting model (S2S-F)}
We hypothesize that the requirement to forecast future time points in the patient's signal would force the encoding LSTM to extract meaningful representations of the past time series.
For this purpose, we design another Seq2Seq-based variant, S2S-F (``F'' for forecasting), where the decoder predicts the future time series instead of reconstructing the past time series in the input.
In this way, the representations still reflect the current patient state but are also optimized to predict the future patient state.
We modify \eqref{eq:s2s-c} and \eqref{eq:s2s-c_obj} to get the decoder function and the loss function for S2S-F:
\begin{equation*}
g^\prime_\mathbf{\phi}(\mathbf{e}_t) = \hat{\mathbf{x}}_{t+1}, \hat{\mathbf{x}}_{t+2}, \ldots,\hat{\mathbf{x}}_{t+T}, \quad\quad\mathcal{L}^\prime(X_t) = (1/T)\sum\nolimits_{\tau=1}^{T} \| \mathbf{x}_{t+\tau} - \hat{\mathbf{x}}_{t+\tau}\|^2.
\end{equation*}

\subsection{Forecasting with attention (S2S-F-A)}
\begin{figure}[!t]
  \begin{center}
  \includegraphics[width=0.75\textwidth,trim={0 6.2cm 0 3.0cm},clip]{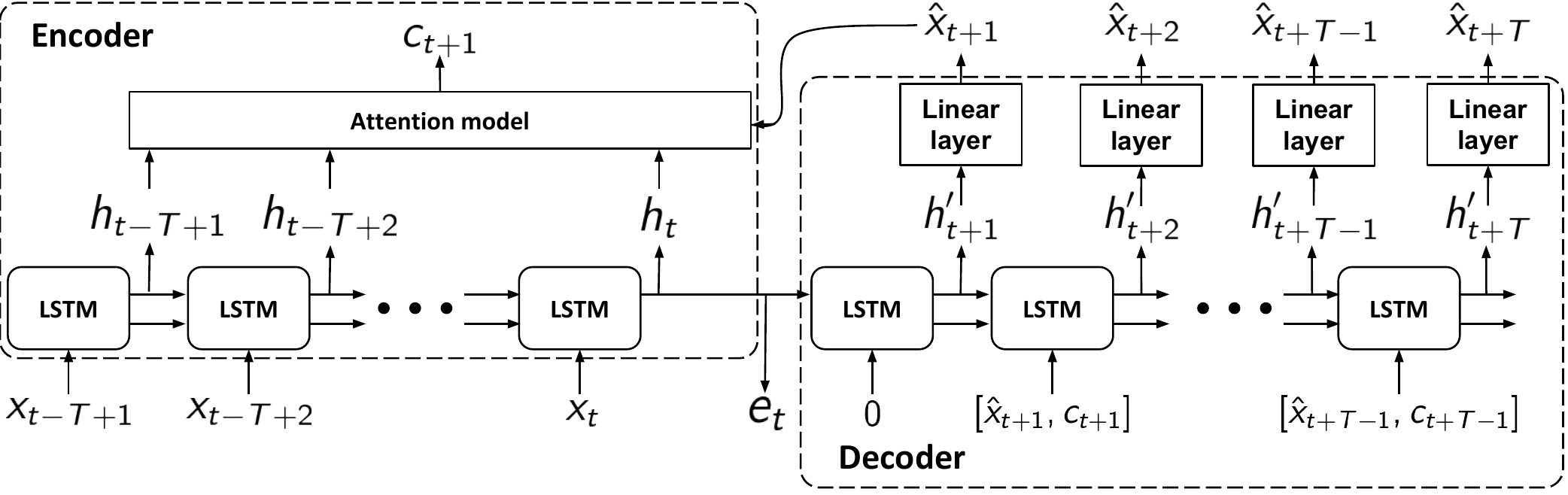}\vspace*{-3.5ex}
  \end{center}
\caption{Sequence-to-Sequence Forecaster with Attention (S2S-F-A).\vspace*{-4.5ex}}
\label{fig:new_Seq2Seq_attention}
\end{figure}

The idea behind applying attention mechanisms to time series forecasting is to enable the decoder to preferentially ``attend'' to specific parts of the input sequence during decoding.
This allows for particularly relevant events (e.g. drastic changes in heart rate), to contribute more to the generation of different points in the output sequence.
Since autoencoding with attention is trivial (an effective attention mechanism would learn to only point to the corresponding input at each time point), we only augment S2S-F with the attention mechanism, calling the architecture S2S-F-A (shown in Figure~\ref{fig:new_Seq2Seq_attention}).

Formally, at time $\tau \in \small\{t+1,\ldots,t+T\small\}$  during decoding, the objective is to produce a context vector $\mathbf{c}_\tau$ which is a weighted combination of the hidden states $\mathbf{h}_j$ of the decoder: $\mathbf{c}_\tau = \sum_{j=1}^{T}\alpha_{\tau j}\mathbf{h}_j. $
The weights $\alpha_{\tau j}$ are softmax-normalized versions of weights $\gamma_{\tau j}$ computed by the attention mechanism $F$, which considers both the current state of the decoder $\mathbf{h'}_\tau$ and each state of the encoder $\mathbf{h}_j$ in turn: $ \gamma_{\tau j} = F(\mathbf{h'}_\tau, \mathbf{h}_j)$ and $\alpha_{\tau j}=\exp({\gamma_{\tau j}})/\sum_{i}\exp({\gamma_{\tau i}})$.
To implement $F$, we use a single-layer perceptron with a {\it tanh} activation function and scalar output, following \cite{luong2015effective}:
$F(\mathbf{h'}_\tau, \mathbf{h}_j) = \mathbf{\beta}^T\text{tanh}(\mathbf{W}_\text{d}\mathbf{h'}_\tau + \mathbf{W}_\text{e}\mathbf{h}_j).$

Each $\alpha_{\tau j}$ reflects the importance of time point $j$ in the input sequence for decoding time point $\tau$ in the output.
The context vector $\mathbf{c}_\tau$ is thus an explicit resummarization of the input data in light of the current decoding task. The context vector is concatenated to the usual input fed to the decoder at $\tau+1$, which is $\hat{\mathbf{x}}_\tau$ (see Figure~\ref{fig:new_Seq2Seq_attention}).

The attention mechanism breaks the "bottleneck" principle of usual Seq2Seq models, and it is not obvious how to choose a self-contained representation. 
Following our practice for S2S-AE and S2S-F, we take the final state of the encoder, $\mathbf{h}_t$ as the representation.
Although we experimented with additionally including context vectors as part of the representation, an interesting finding was that simply taking $\mathbf{h}_t$ was sufficient in the prediction of downstream tasks. Table~\ref{table:models} summarizes the characteristics of the unsupervised representation models we analyze.

%% file: experiments.tex
\vspace{-0.2cm}
\section{Experiments and results}
\vspace{-0.2cm}

\paragraph{Data}
The eICU Collaborative Research Database v1.2~\citep{goldberger00_physiobank_physitoolkit_physionet} was used for all experiments described in this paper.
94 time series variables including periodic and aperiodic vital signs and irregularly measured lab tests were extracted.
The data was resampled to be hourly, with implausible data rejection and imputation performed online; see Appendix~\ref{subsec:data} for more details. Overall, the dataset consists of 20,878 patients with 72-720 hours of history, extending from ICU admission to dispatch.
We use a window size of 12 hours (i.e. 12 time points) and representation dimension 94.

\paragraph{Reconstructing past and predicting future} 
We aim to evaluate the ability of representations to reconstruct past and future data.
Some representations are obtained from models optimized to reconstruct past data (PCA, AE and S2S-AE), while others from models optimized to predict future data (S2S-F and S2S-F-A).
To produce a fair comparison independent of a specific decoder, we use the representations themselves as input features to a 1-layer LSTM trained to either reconstruct the past 12 hours, or predict the next 12. The performance for each set of representations are shown in Table~\ref{tbl:recon_result}, evaluated using mean-squared error (MSE).
Not surprisingly, representations from forecaster models perform better in future prediction and the attention mechanism further improves performance.
However, the extent to which attention helps is surprising.
\begin{table}[ht!]
  \vspace*{-3ex}
  \centering
  \caption{Performance of representations used as input features to a 1-layer LSTM trained to either reconstruct the past 12 hours, or predict the next 12 hours. (The best results are in bold and second are best marked with *)}
  \tiny\begin{tabular}{l c c c c c}
    \midrule \vspace*{-0.7ex}
    {\bf MSE} & {\bf PCA} & {\bf AE} & {\bf S2S-AE} & {\bf S2S-F} & {\bf S2S-F-A} \\ \midrule \vspace*{-0.7ex}
    {\bf Reconstruction} & $0.0743 \pm 0.002$ & $\mathbf{0.0403 \pm 0.001}$ & $0.0505 \pm 0.001$ & $0.0500 \pm 0.001$ & ${0.0474 \pm 0.003}^*$ \\ \midrule \vspace*{-0.7ex}
    {\bf Prediction} & $0.149 \pm 0.003$ & ${0.114 \pm 0.002}^*$ & $0.121 \pm 0.003$ & $0.119 \pm 0.003$ & $\mathbf{0.0982 \pm 0.003}$ \\ \midrule
    \end{tabular}\vspace*{-3ex}
  \label{tbl:recon_result}
\end{table}

\paragraph{Predicting mortality and discharge status within the next 24 hours}

Besides evaluating the ability of representations in past/future signal prediction, we are also interested in whether we can use them to predict future clinical events.
Here we focus on predicting whether patients will be discharged from the ICU in a stable state (``24h Discharge''), or die within the next 24 hours (``24h Mortality'').
We trained 1-layer LSTM classifiers (LSTM-1) using representations as input to predict these two events and report the area under ROC curve (AUROC) and the area under precision-recall curve (AUPRC) in Table~\ref{tbl:lstm_result}.
In addition, we also include the performance of a 3-layer LSTM classifier (LSTM-3), a ``deeper'' model, trained on the original input signals as a baseline. 

\begin{table}[!h]
\vspace*{-3ex}
  \caption{Prediction of discharge/mortality status within the next 24 hours using unsupervised representations or raw signals. The prevalence of the discharge and mortality positive labels is 0.197 and 0.021 respectively. (The best results are in bold and second are best marked with *) }
  \label{tbl:lstm_result}
  \centering
  \tiny\begin{tabular}{r r c c c c}
     ~ & ~& \multicolumn{2}{c}{\bf 24h Discharge} & \multicolumn{2}{c}{\bf 24h Mortality} \\ \cmidrule{3-6} \vspace*{-0.7ex}
     ~ & ~& AUPRC & AUROC & AUPRC & AUROC \\ \midrule \vspace*{-0.7ex}
     \multirow{5}{*}{LSTM-1 +} & PCA rep. & $ 0.436 \pm 0.01 $ & $ 0.811 \pm 0.004 $ & $ 0.0975 \pm 0.007 $ & $ 0.78 \pm 0.007 $ \\ \cmidrule{2-6} \vspace*{-0.7ex}
     & AE rep.  & $ 0.471 \pm 0.005 $ & $ 0.824 \pm 0.002 $ & $ \mathbf{0.203 \pm 0.01} $ & $ 0.889 \pm 0.008^* $  \\\cmidrule{2-6} \vspace*{-0.7ex}
     & S2S-AE rep. & $ 0.474 \pm 0.006 $ & $ 0.824 \pm 0.003 $ & $ 0.196 \pm 0.02 $ & $ 0.887 \pm 0.004 $  \\ \cmidrule{2-6} \vspace*{-0.7ex}
     & S2S-F rep. & $ 0.477 \pm 0.006 ^* $ & $ 0.825 \pm 0.003^* $  & $ 0.193 \pm 0.01 $ & $ 0.886 \pm 0.005 $ \\ \cmidrule{2-6} \vspace*{-0.7ex}
     & S2S-F-A rep. & $ \mathbf{0.48 \pm 0.007} $ & $ 0.825 \pm 0.003^* $ & $ 0.201 \pm 0.01^* $ & $ \mathbf{0.89 \pm 0.009} $ \\  \midrule  \vspace*{-0.7ex}
     LSTM-3 + & raw signals & $ 0.438 \pm 0.01 $ & $\mathbf{ 0.834 \pm 0.002} $ & $ 0.181 \pm 0.01 $ & $ \mathbf{0.89 \pm 0.01} $ \\ \midrule
  \end{tabular}\vspace*{-3ex}
\end{table}

\paragraph{Improved performance in limited data setting}

Here we evaluate how unsupervised representations help boost prediction performance in the limited labeled data scenario.

\begin{figure}[!ht]
\vspace*{-1.5ex}
\centering
        \includegraphics[width=0.9\textwidth,trim={0.25cm 0 0.1cm 0.2cm},clip]{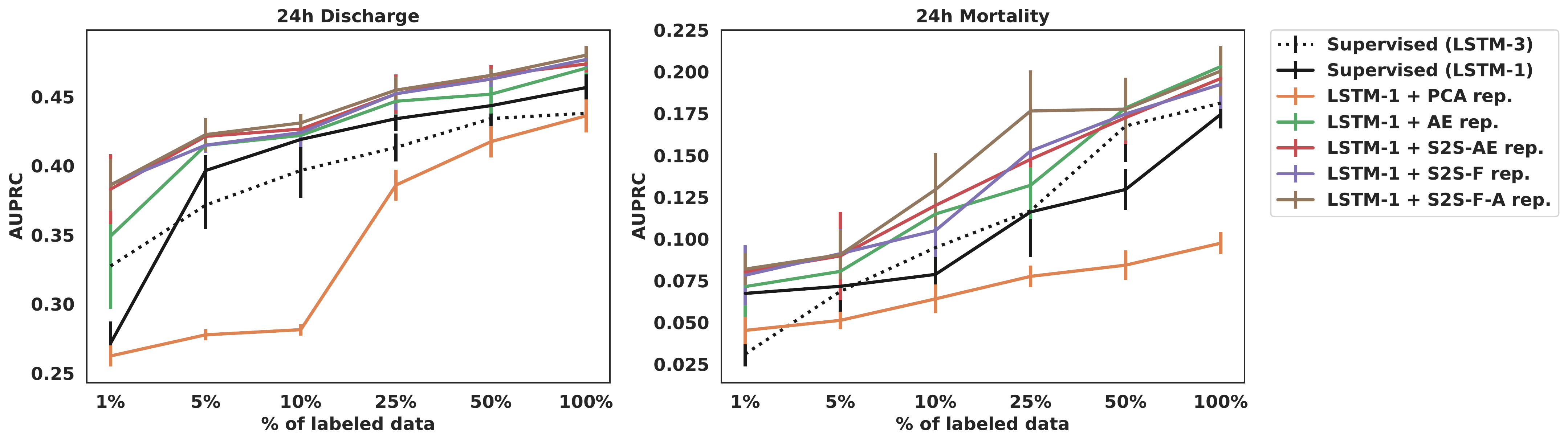}\vspace*{-2ex}
        \caption{24h discharge and mortality prediction performance of LSTM-1 using unsupervised representations, as well as supervised learning with LSTM-1 and LSTM-3. (There are only 75 labeled patients for training in the 1\% labeled data setting.)}\vspace*{-1.5ex}
        \label{fig:limited_data_lr_lstm}
\end{figure}

We simulate this setting by reducing the quantity of \emph{labeled} data available for the classification problems described in the previous section, with as few as $1\%$ (N = 75 patients) training examples. The results under this varying data scarcity are shown in Figure~\ref{fig:limited_data_lr_lstm}, for the different representation-learning approaches.
We also include the prediction performance of classifiers, namely LSTM-1 and LSTM-3, trained in an end-to-end supervised fashion on the available labeled data, as baselines.

We observe from Figure \ref{fig:limited_data_lr_lstm} that when labels are scarce, the model trained using time-series representations as input features outperforms the end-to-end supervised model, confirming the benefit of unsupervised representation learning in limited data settings. Even when we use all labeled samples at our disposal to train a more complex classifier, the best unsupervised representations still lead to a better performance than supervised representations. For all models, however, performance does not saturate when increasing the training set size, which indicates that the entire regime examined here is the data scarcity regime. Given more data, the purely supervised models might eventually surpass the ones using learned representations.

%% file: conclusion.tex
\vspace{-0.2cm}
\section{Conclusion}
\vspace{-0.2cm}
We have studied the performance of several methods for learning unsupervised representations of patient time series, and proposed a new architecture, S2S-F-A, which is optimized for forecasting using an attention mechanism.
We empirically showed that in scenarios where labeled medical time series data is scarce, training classifiers on unsupervised representations provides performance gains over end-to-end supervised learning using raw input signals, thus making effective use of information available in a separate, unlabeled training set.
The proposed model, explored for the first time in the context of unsupervised patient representation learning, produces representations with the highest performance in future signal prediction and clinical outcome prediction, exceeding several baselines.

%% file: appendix.tex
\begin{appendix}
\renewcommand\thefigure{\thesection\arabic{figure}}    
\setcounter{figure}{0}
\renewcommand{\thetable}{\thesection\arabic{table}}
\setcounter{table}{0}
\section{Appendix}

\subsection{Data}\label{subsec:data}

The eICU Collaborative Research Database v1.2 \citep{goldberger00_physiobank_physitoolkit_physionet} was used for all experiments described in this paper.
94 time series variables (shown in Table~\ref{tbl:variables}) including periodic and aperiodic vital signs and irregularly measured lab tests were extracted from the raw database. A variable 
was included in our analysis if at least 10\% of patients in the cohort had at least one record for this variable. As preprocessing, the raw data was
resampled to a regular time-grid format with an interval size of 60 minutes, extending from admission to the ICU to dispatch from the unit.
During computation of the time grid, rejection of implausible data and imputation were performed with an online algorithm. An observation was rejected if
it is a statistical outlier with respect to pre-computed 5th/95th dataset percentiles. Values on the regular time grid were imputed using
a combination of forward filling, personalized history mean filling and population median filling. Forward filling was used if the last value
was recorded no earlier than 1 hour (periodic vital signs), 5 hours (aperiodic vital signs) or 1 day (lab tests) prior to the grid point, respectively.
Otherwise, if there have been previous observations of that variable, the mean of all such observations was used to fill in the time grid point. If there were
no observations in a patient's history, the grid value was filled with the population median for that variable.

Overall, the dataset consists of 20878 patients with 72-240 hours of history.
\begin{table}[!h]
  \caption{List of 94 selected variables.}
  \label{tbl:variables}
  \scriptsize\begin{tabular}{l p{0.8\textwidth}}
    \midrule
    \textbf{eICU Table} & \textbf{Variables} \\ \midrule
    vitalPeriodic & cvp, heartrate, respiration, sao2, st1, st2, st3, systemicdiastolic, systemicmean, systemicsystolic, temperature\\\midrule
    vitalAperiodic & noninvasivediastolic, noninvasivemean, noninvasivesystolic\\ \midrule
    Lab & -bands, -basos, -eos, -lymphs, -monos, -polys, ALT (SGPT), AST (SGOT), BNP, BUN, Base Deficit, Base Excess, CPK, CPK-MB, CPK-MB index, Carboxyhemoglobin, Fe, Ferritin, FiO2, HCO3, HDL, Hct, Hgb, LDL, LPM O2, MCH, MCHC, MCV, MPV, Methemoglobin, O2 Content, O2 Sat (\%), PT, PT - INR, PTT, RBC, RDW, Respiratory Rate, TIBC, TSH, TV, Total CO2, Vancomycin - trough, Vent Rate, Vitamin B12, WBC x 1000, WBC's in urine, albumin, alkaline phos., ammonia, anion gap, bedside glucose, bicarbonate, calcium, chloride, creatinine, direct bilirubin, fibrinogen, glucose, ionized calcium, lactate, lipase, magnesium, pH, paCO2, paO2, peep, phosphate, platelets x 1000, potassium, sodium, temporature, total bilirubin, total cholesterol, total protein, triglycerides, troponin - I, troponin - T, urinary sodium, urinary specific gravity \\ \midrule
  \end{tabular}
 
\end{table}

\subsubsection{Cohort selection}

Among the >200,000 ICU stays available in the dataset, we 
included only patients with one stay, such that
data splits do not have to be stratified with respect to patient
ID. In the second filtering step, ICU stays shorter than 3 days or
longer than 10 days were excluded. The filtering yielded a set of 20878
patients/ICU stays.

\subsubsection{Data splits}

From the pre-filtered dataset we created 5 replicates of
random partitions into train, validation and 2 test sets, with respect
to patients, i.e. the entire data of a patient was contained
in exactly one of the 4 sets. Size ratios of 40:40:10:10 for
train/validation/test1/test2 sets were used. The training set was used to
train the representations, the validation set was used to tune free
hyperparameters of the representation method (if any). The classifiers were trained on the
patient representations obtained from the validation set, optimized its
hyperparameters on the representations from the first test set, and its
predictive performance was evaluated on the unseen representations from the
second test set. 5 independent experiments have been performed on the
replicates.

\subsection{Representation learning}
For each representation learning method, representations were extracted from the
training set. Feature columns were standard-scaled (subtracting mean /
dividing by the standard deviation) before training the models to obtain representations. The validation set was used to implement an early stopping
heuristic for the training process, in the case of the deep learning
models. At this point, all trained representations were saved to
disk. For the deep learning models, we used grid search to find the best set of hyper-parameters.

For basic autoencoders, we train with a mini-batch of 512 
randomly sampled records, and for the recurrent
autoencoders we train with a mini-batch of 4 patients
with full history. We use early
stopping based on the validation set loss to avoid overfitting,
i.e. we stop training if we observe that validation set loss is
non-decreasing for 10 consecutive epochs.  We additionally use the
validation set to perform hyperparameter optimization over the optimal
learning rate and activation functions.

\subsection{Representation evaluation}
For evaluating the future signal and task prediction performance, representations of
the first 12 hours of a patient recording were excluded. In this way
the results are not affected by the model-specific ways of handling
incomplete histories, which occur at the beginning of the patient
stay.

\subsection{Model complexity}
Table~\ref{table:models} shows the traits of the unsupervised learning models used in the paper.
An advantage of Seq2Seq-based models is that the number of parameters they use does not depend on the length of the input time series to be compressed.

\begin{table}[!ht]
    \caption{Comparison of used unsupervised representation learning models. $T$ refers to the length of the time series to be encoded (12 in our experiments), $d$ is the dimension of the input data, and $m$ is the dimension of the hidden state of the LSTM in the S2S-based models, which is the same as the representation dimension.}
    \label{table:models}
  \centering
  \scriptsize  \begin{tabular}{ l c  c c c c}\midrule
        name & nonlinear & temporal & decoder output & attention & number of parameters \\\midrule
        PCA & & & past & & $\mathcal{O}(T m d)$ \\\midrule
        AE & \checkmark & & past & & $\mathcal{O}(T m d)$ \\\midrule
        S2S-AE & \checkmark & \checkmark & past & & $\mathcal{O}(m^2 + md)$ \\\midrule
        S2S-F & \checkmark & \checkmark & future & & $\mathcal{O}(m^2 + md)$ \\\midrule
        S2S-F-A & \checkmark & \checkmark & future & \checkmark & $\mathcal{O}(m^2 + md)$\\\midrule
    \end{tabular}
x\end{table}

\subsection{Impact of representation dimension}
In this section we investigate the relationship between the dimensionality of representations and their performance across tasks.
In the previously described experiments, we used a representation dimension of $m = 94$, implying a compression factor of 12 (as the windows consist of 12 hourly measurements of 94 variables).
Here we vary the value of $m$ to explore how much compression is possible while retaining prediction performance.

Table \ref{tbl:low_dim_embed_raw_comparison} shows the AUROC values using S2S-F-A representations for prediction.
Compared with the AUROC scores corresponding to using raw features in Table \ref{tbl:lstm_result}, even the S2S-F-A representations with very low dimension still obtain reasonable performance.
\begin{table}[!h]
  \caption{AUROC scores of predictions using LSTM-1 classifiers on S2S-F-A representations with different dimensions.}
  \label{tbl:low_dim_embed_raw_comparison}
  \centering
  \scriptsize
    \begin{tabular}{r c c c}
    \midrule
    \textbf{AUROC} & {\bf S2S-F-A (m=2)} & {\bf S2S-F-A (m=50)} & {\bf S2S-F-A (m=94)}\\ \midrule
    {\bf 24h Discharge} & $ 0.7985 \pm 0.003 $ & $ 0.8028 \pm 0.004 $ & $ 0.825 \pm 0.003 $ \\ \midrule
    {\bf 24h Mortality} &  $ 0.8398 \pm 0.01 $ & $ 0.8696 \pm 0.006 $ &  $0.89 \pm 0.009 $  \\  \midrule
  \end{tabular}\vspace*{-3ex}
\end{table}
\end{appendix}